\documentclass[runningheads]{llncs}
\usepackage{graphicx}
\usepackage{amsmath,amssymb} 
\usepackage{color}\usepackage[width=122mm,left=12mm,paperwidth=146mm,height=193mm,top=12mm,paperheight=217mm]{geometry}

\usepackage{enumitem}
\usepackage{tensor}
\usepackage[table,xcdraw]{xcolor}
\usepackage{stackengine}
\newcolumntype{L}{>{\centering\arraybackslash}m{1.5cm}}

\begin{document}
\pagestyle{headings}
\mainmatter

\title{Weakly Supervised Localization using Deep Feature Maps} 

\titlerunning{Weakly Supervised Localization using Deep Feature Maps}

\authorrunning{Bency \and Kwon  \and Lee \and Karthikeyan \and Manjunath}

\author{Archith J. Bency \inst{1} \and Heesung Kwon \inst{2} \and Hyungtae Lee \inst{2} \and S. Karthikeyan \inst{1} \and \\ B. S. Manjunath \inst{1}}
\institute{ University of California, Santa Barbara, CA, USA \and  Army Research Laboratory, Adelphi, MD}

\maketitle

\begin{abstract}
   Object localization is an important computer vision problem with a variety of applications. The lack of large scale object-level annotations and the relative abundance of image-level labels makes a compelling case for weak supervision in the object localization task. Deep Convolutional Neural Networks are a class of state-of-the-art methods for the related problem of object recognition. In this paper, we describe a novel object localization algorithm  which uses classification networks trained on only image labels. This weakly supervised method leverages local spatial and semantic patterns captured in the convolutional layers of classification networks. We propose an efficient beam search based approach to detect and localize multiple objects in images. The proposed method significantly outperforms the state-of-the-art in standard object localization data-sets with a 8 point increase in mAP scores.

\keywords{Weakly Supervised methods, Object localization, Deep Convolutional Networks}
\end{abstract}

\section{Introduction}
Given an image, an object localization method aims to recognize and locate interesting objects within the image. The ability to localize objects in images and videos efficiently and accurately opens up a lot of applications like automated vehicular systems, searching online shopping catalogues, home and health-care automation among others. Objects can occur in images in varying conditions of occlusion, illumination, scale, pose and context. These variations make object detection a challenging problems in the field of computer vision.

The current state of the art in object detection includes methods which involve `strong' supervision. In the context of  object detection , strong supervision entails annotating  localization and pose information about present objects of interest. Generating such rich annotations is a time-consuming process and is expensive to perform over large data-sets. Weak supervision lends itself to large-scale object detection for data-sets where only image-level labels are available. Effective localization under weak supervision enables extensions to new object classes and modalities without human-generated object bounding box annotations. Also, such methods enable generation of inexpensive training data for training object detectors with strong supervision.

Deep Convolutional Neural Networks  (CNNs) \cite{szegedy2015going}, \cite{krizhevsky2012imagenet} have created new benchmarks in the object recognition challenge \cite{deng2009imagenet}. CNNs for object recognition are trained using image-level labels to predict the presence of objects of interest in new test images.  A common paradigm in analyzing CNNs has emerged where the convolutional layers are considered as data-driven feature extractors and the subsequent fully-connected layers constitute hyperplanes which delineate object categories in the learnt feature space. Non-linearities through Rectified Linear Units (ReLU) and sigmoidal transfer functions have helped to learn complex mapping functions which relate images to labels.  The convolutional layers encode both semantic and spatial information extracted from training data. This information is represented by  activations from the convolutional units in the network which are commonly termed as Feature Maps.

\begin{figure}
\begin{center}
\includegraphics[width=\columnwidth]{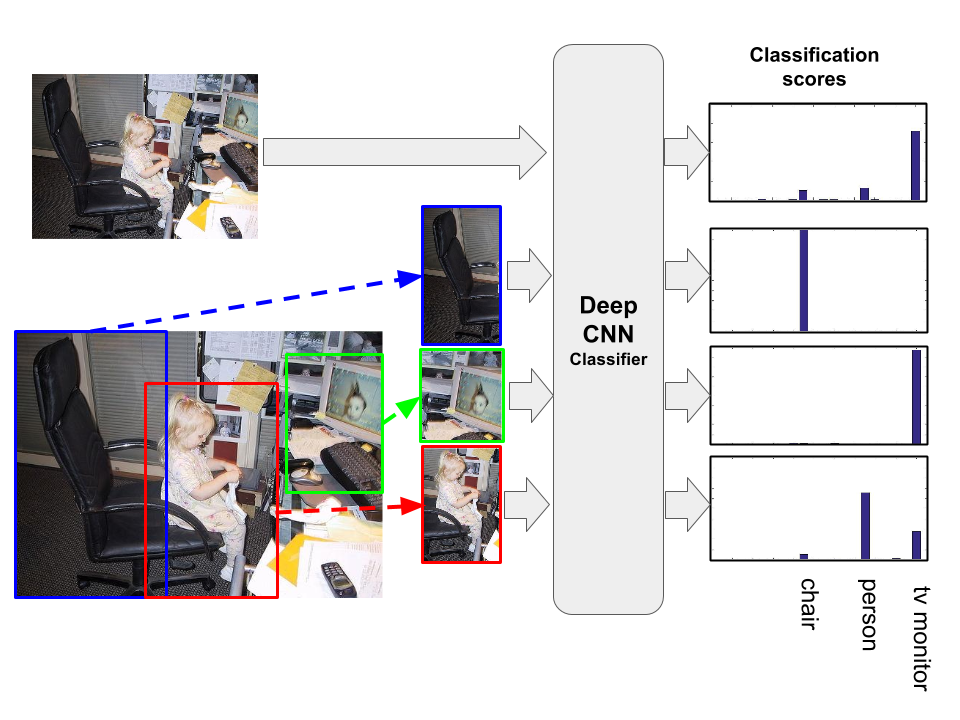}
\end{center}
   \caption{When localizations centered around objects of interest are classified by Deep CNNs, the corresponding object classes are assigned high scores.}
\label{fig:localization_bias}
\end{figure}

In this paper, we present a method that exploits  correlation between semantic information present in Feature Maps and  localization of an object of interest within an  image. An example of such correlation can be seen in Figure $\ref{fig:localization_bias}$. Note that crudely localized image-patches with the objects of classes,`chair', `person' and `tv monitor', generate high classification scores for the corresponding classes. This suggests that one can coarsely localize objects solely by image classification scores in this context.

CNN based classifiers are trained for the task of image recognition on large image classification data-sets \cite{deng2009imagenet}, \cite{pascal-voc-2012}, \cite{pascal-voc-2007}. The learnt convolutional filters compute spatially localized activations across layers for a given test image \cite{mahendran2014understanding}. We examine  the activation values in the outermost convolutional layer and propose localization candidates (or bounding boxes) which maximize classification scores for a class of interest. Class scores vary across localization candidates because of the aforementioned local nature of the convolutional filters. We then progressively explore smaller and smaller regions of interest till a point is reached where the classifier is no longer able to discriminate amongst the classes of interest. The localization candidates are organized in a search tree, the root node being represented by the entire test image. As we traverse from the root node towards the leaf nodes, we consider finer regions of interest. To approximate the search for optimal localization candidates, we adopt a beam search strategy where the number of candidate bounding boxes are restricted as we progress to finer localizations. This strategy enables efficient localization of multiple objects of multiple classes in images. We  outperform the state-of-the-art in localization accuracy by a significant margin of up to 8 mAP on two standard data-sets with complex scenes, PASCAL VOC 2012 \cite{pascal-voc-2012} and the much larger MS COCO \cite{lin2014microsoft}.

The main contributions of this paper are:

\begin{itemize}
    \item{We present a method that tackles the problem of object localization for images in a weakly supervised setting using deep convolutional neural networks trained for the simpler task of image-level classification.}

    \item{We propose a method where the correlation between spatial and semantic information in the convolutional layers and localization of objects in images  is used explicitly for the localization problem.}
\end{itemize}

\section{Related Work}\label{section:relatedWorks}

The task of object detection is one of the fundamental problems in computer vision with wide applicability. Variability of object appearance in images makes object detection and localization a very challenging task and thus has attracted a large body of work. Surveys of the state-of-the-art are provided in \cite{Zhang:2013:OCD:2522968.2522978}, \cite{survey1}.

A large selection of relevant work are trained in the strong supervision paradigm with detailed  annotated ground truth in the form of  bounding boxes \cite{viola2001rapid}, \cite{felzenszwalb2010object}, object masks \cite{brox2011object}, \cite{kim2012shape}, \cite{hariharan2014simultaneous} and 3D object appearance cues \cite{glasner2011viewpoint},\cite{shrivastava2013building}. The requirement of rich annotations curb the application of these methods in data-sets and modalities where training data is limited to weaker forms of labeling. Weak supervision for object detection tries to work around this limitation by learning localization cues from large collection of data with in-expensive annotations.

Large data-sets like Imagenet \cite{deng2009imagenet} and MS COCO  are available with image-level labels. There has been significant work in this direction for object localization and segmentation \cite{galleguillos2008weakly}, \cite{chum2007exemplar}, \cite{hartmann2012weakly}, \cite{blaschko2010simultaneous}, \cite{deselaers2010localizing}, \cite{pourian2015iccv}. Apart from image-level labels, other kinds of weak supervision include using eye-tracking data \cite{papadopoulos2014training}, \cite{karthikeyaneye}. 

Deep convolutional neural networks (CNN) have seen a surge of attention from the computer vision community in the recent years. New benchmarks have been created in diverse tasks such as image classification and recognition \cite{simonyan2014very}, \cite{krizhevsky2012imagenet}, \cite{szegedy2015going}, \cite{chatfield2014return}, object detection \cite{girshick2014rich},
\cite{sermanet2013overfeat}, \cite{zhu2015segdeepm}, \cite{Zhang_2015_CVPR}, \cite{Ouyang_2015_CVPR} and object segmentation \cite{long2015fully}, \cite{chen14semantic}, \cite{noh2015learning} among others by methods building on deep convolutional network architectures. These networks perform tasks using feature representations learnt from training data instead of traditional hand-engineered features \cite{dalal2005histograms}, \cite{felzenszwalb2010object}, \cite{murphy2006object}.
Typical algorithms of this paradigm perform inference over the last layer of the network. There have been recent works \cite{hariharan2014hypercolumns}, \cite{dai2015convolutional}, \cite{he2015spatial} which exploit semantic information encoded in convolutional feature map activations for semantic segmentation and object detection. A prerequisite for these CNN-based  algorithms is strong supervision with systems focused on detection requiring location masks or object bounding boxes for training. \cite{scenecnn_iclr15} studies the presence of object detector characteristics in image-classification CNNs, but does not provide a computational method to carry out object detection.

 Oquab et.al. \cite{oquabobject} has proposed a weakly supervised object localization system which learns from training samples with objects in composite scenes by explicitly searching over candidate object locations and scales during the training phase. While this method performs well on data-sets with complex scenes, the extent of localization is limited with respect to estimating one point  in the test image. The extent of the object is not  estimated and detecting multiple instances of the same object class is not considered. In our proposed approach, we estimate both the location and extent of objects and are capable of estimating multiple instances of objects in the test image. Also, we use pre-existing classification networks for localization where as \cite{oquabobject} proposes training custom adaptation layers.

\section{Weakly Supervised Object Localization }\label{section:WeakSup}

\begin{figure*}
\begin{center}
\includegraphics[width=\columnwidth]{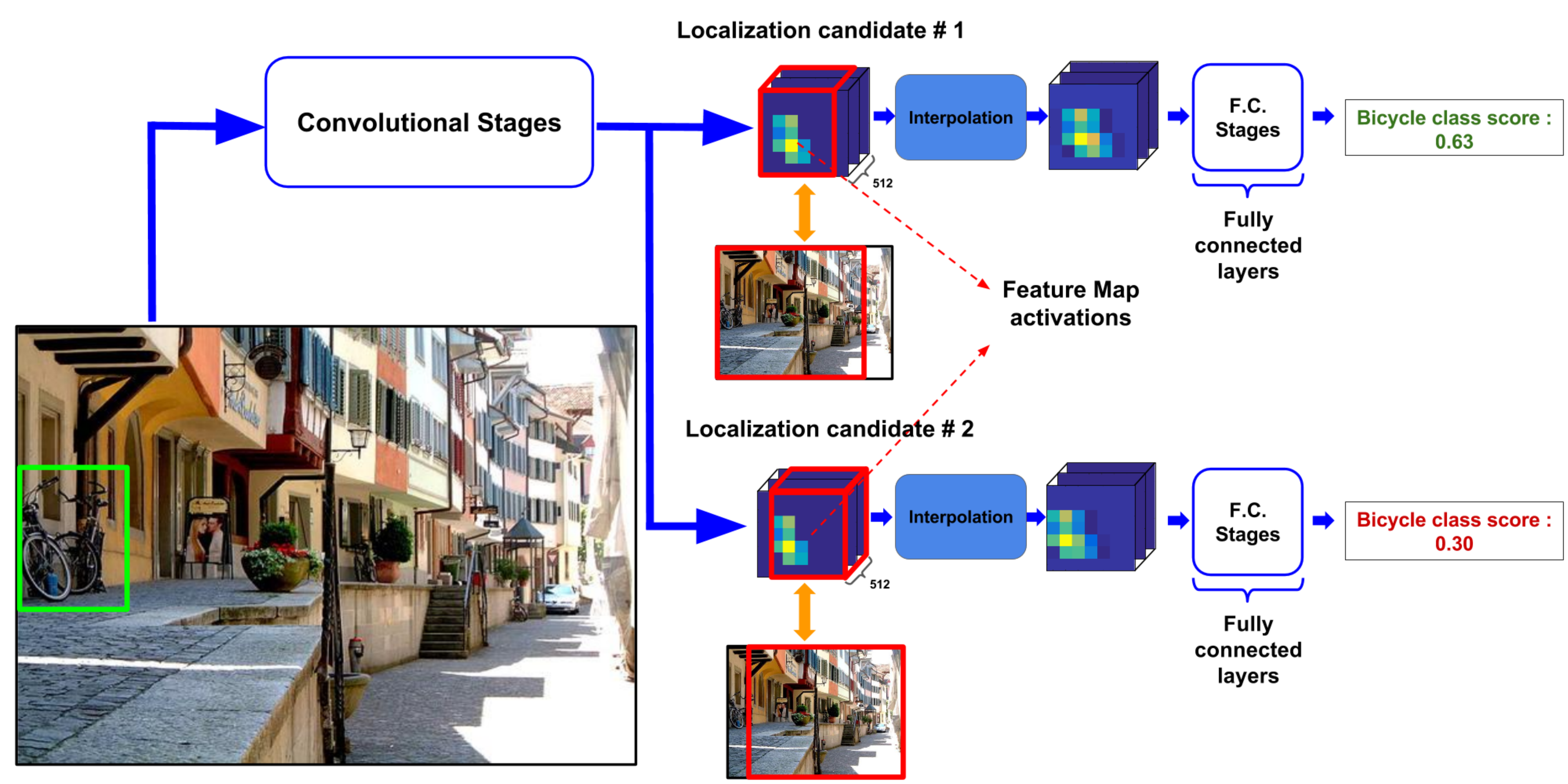}
\end{center}
   \caption{An illustration of how two different localization candidates are compared in the localization process. Candidate \# 1 scores higher for the bicycle class than candidate \# 2. The first candidate is further iterated upon to achieve finer localization. The green box in the left image denotes ground-truth location of the bicycle object.}
\label{fig:activation_masking_cropped}
\end{figure*}

\subsection{Overview of the method}
We aim to localize and recognize objects in images using CNNs trained for classification. There are two distinct phases. The first phase consists of learning image-level recognition from training image sets using existing Deep CNN architectures. We use the popular Alexnet \cite{krizhevsky2012imagenet} and VGG-16 \cite{simonyan2014very} networks for our experiments. The next phase involves generating localization candidates in the form of bounding boxes for object classes of interest. These candidates are generated from a spatial grid corresponding to the final convolutional layer of the network and are organized in a search tree. We carry out a beam-search based exploration of these candidates with the image classifier scoring the candidates and reach at a set of final localization candidates for each class of interest.

\subsection{Network architecture and training}\label{subsection:network}
The Alexnet network  has five convolutional layers with associated rectification and  pooling layers $C_1, C_2,\ldots, C_5$, along with three fully connected layers $F_6, F_7, F_8$ with $M_6 = \sigma(W_6 M_5 + B_6)$, $M_7 = \sigma(W_7 M_6 + B_7)$ and $M_8 = \gamma(W_8 M_7 + B_8)$. $W_n, B_n$ are learn-able parameters  for the $n$-th layer, $M_n$ is the output of the $n$-th layer. $\sigma(\textbf{X}) = \max(\textbf{0}, \textbf{X})$ is the rectification function and $\gamma(\textbf{X}) = [e^{\textbf{X}[i]}/\Sigma_{j}e^{\textbf{X}[j]}]$ is the softmax function. Of particular interest to us is the output of the last convolutional layer $C_5$, $M_5$ which we will refer to subsequent sections.

We learn the network parameters through stochastic gradient descent and back-propagation of learning loss error \cite{rumelhart1988learning} from the classification layer back through the fully connected and convolutional layers. Keeping in mind that objects of multiple classes can be present in the same training image, we use the cross entropy loss function to model error loss $J$ between ground truth class probabilities $\{{p_{k}}\}$ and predicted class probabilities $\{{\hat{p}_k}\}$, where $k \in \{ 0, 1, ... , K-1\}$ indexes the class labels.

\begin{equation}
J = -\frac{1}{K}\sum_{k=0}^{K-1}[p_{k} \log\hat{p}_{k} + (1-p_{k}) \log(1-\hat{p}_{k})]
\end{equation}

As specified in \cite{Oquab14}, we remove $F_8$ and add two additional fully connected adaptation layers $F_{a}, F_{b}$. Similar to the Alexnet network, the ouput of these layers are computed as $M_a = \sigma(W_a M_7 + B_a)$ and $M_b = \gamma(W_b M_a + B_b)$. In order to assess the effectiveness of the proposed method for localization, these additional layers are added to facilitate re-training of the network from the Imagenet data-set to the Pascal VOC or MS COCO object detection data-sets. We  initialize network parameters to values trained on the Imagenet data-set and fine-tune them \cite{karayev2013recognizing} to adapt onto a target data-set. This is achieved by setting the learning rate parameter  for the last layer weights to a higher value relative to earlier layer weights. An illustration of the network architecture is presented in Figure 2 of \cite{Oquab14}.

We train the augmented network on labeled samples from the target data-set. The trained network produces class scores at the final layer which are treated as probability estimates of the presence of a class in the test image.

The VGG-16 network, being similar to the Alexnet network, has thirteen convolutional layers $C_1, C_2, C_3, .... C_{13}$ with associated rectification and  pooling layers, along with three fully connected layers $F_6, F_7, F_8$. Similar to the Alexnet network, the feature map $M_{13}$ is of special interest to us.
The increased number of layers and associated learnable parameters provides an improved image recognition performance when compared to the Alexnet network. The improvement however comes at the cost of increased GPU memory (442 MB vs 735 MB) and computations (6 milliseconds  vs 26 milliseconds for classifying an image).

In addition to using image-labels to train the deep CNNs, we also use label co-occurrence information to improve classification. Some classes tend to occur together frequently. For example, people and motorbikes or people and chairs tend to share training samples. We treat the class scores from the classifier as unary scores and combine them with the likelihood of co-existence of multiple objects of different classes in the same object. We model the co-existence likelihood by building a co-occurrence matrix for class labels from the training data-set.  For the class $b_i$,

\begin{align}
s_{comb}(b_i) &= s_{unary}(b_i) + \alpha \sum_{i \neq j}s_{pair}(b_i|b_j) \\ \label{eq:rescoring}
s_{pair}(b_i| b_j) &= p_{pair}(b_i|b_j)s_{unary}(b_j) \\
p_{pair}(b_i|b_j) &= \frac{|b_i \cap b_j|}{|b_j|}
\end{align}

\noindent where $s_{unary}$ is the initial classification score for the test image, $s_{pair}$ is the pairwise score, $|b_i \cap b_j|$ denotes the number of training samples containing the labels $b_i$ and $b_j$ and $s_{comb}$ is the combined score which we use to re-score the classes for the test image. The parameter $\alpha$ denotes the importance given for pair-wise information in re-scoring. An optimal value is derived by testing over a randomly sampled validation sub-set from the training set.

\subsection{Localization}

In deep CNNs trained for classification, feature map activation values are the result of repeated localized convolutions, rectification (or other non-linear operations) and spatial pooling. Hence the structure of the network inherently provides a receptive field for each activation on the input image. The foot-print region becomes progressively coarser as we go deeper in the layers towards the fully connected layers. In a first attempt, we explore ways to exploit the spatial information encoded in the last convolutional layer for object localization.

Also, standard state-of-the-art object recognition data-sets (for e.g. Imagenet) typically have the object of interest represented in the middle of training samples. This gives rise to a bias in the classifier performance where more centered an object is in the input image, higher the corresponding class score becomes. An example is illustrated in Figure $\ref{fig:localization_bias}$.  The correlation between the location of objects and class scores has been observed in other works \cite{Oquab14}, \cite{girshick2014rich}.

\begin{figure*}
\begin{center}
\includegraphics[width=\columnwidth]{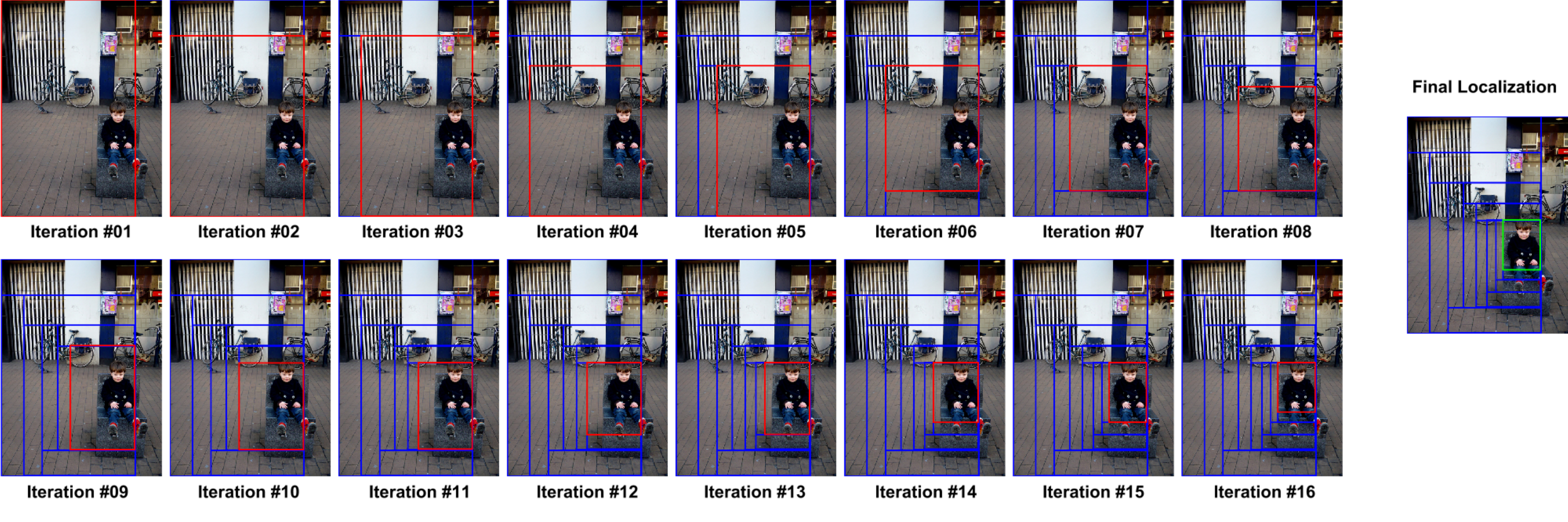}
\end{center}
   \caption{A visual result of the proposed localization strategy on an image. The class scores for `person' category are used to progressively localize the object of interest. Blue rectangles represent localization candidates considered in previous iterations and red rectangles represent current candidates.}
\label{fig:localization_iterations}
\end{figure*}

A naive approach to exploit the correlation would be to carry out a multi-scale sliding window sampling of sub-images from the test sample and spatially pool the classifier scores to generate a heat map of possible object locations for a given object class $C$. The number of sub-images required for effective localization can be in the order of thousands. Although powerful hardware like GPUs have brought image recognition CNNs  into the domain of real-time methods, processing a large number of windows for every test sample is prohibitively expensive. A class of object detection methods \cite{girshick2014rich} try to reduce the number of candidate windows by using object region proposal methods \cite{uijlings2013selective}, \cite{alexe2012measuring}. Time taken to detect objects in each image using these methods still range in tens of seconds when using powerful GPUs.

For a more computationally efficient approach, we take advantage of the spatial and semantic information encoded in the final convolutional feature maps to guide the search process. We refer to the maps as $M_5$ for Alexnet and $M_{13}$ for VGG-16 in the section $\ref{subsection:network}$. For a general CNN network, the final convolutional layer is of size $L\times L \times T$ which means there are $T$ feature maps of size $L \times L$. For the Alexnet and VGG-16 networks, the feature maps are of size  $6\times6\times256$ and $7\times7\times512$ respectively.

Given a test image $I$, we forward propagate the layer responses for the image up-to the final convolutional layer $C_{last}$ and generate the feature map activations $M_{last}$. We generate localization candidates which are sub-grids of the $L \times L$ grid. In concrete terms, these candidates are parametrised as  boxes $b_i = [x_i, y_i, w_i, h_i]$ for $i = 1,2,\ldots,B$ where $x$, $y$, $w$ and $h$ represent the coordinates of the upper-left corner, width and height and $B$ is the total number of possible sub-grids.
For each localization candidate, we sample the feature map activations contained within the  corresponding boxes and interpolate them over the entire $L \times L$ grid. This is done independently over all $T$ feature maps. For the box $b_i$,

\begin{align*}
\hat{M}_{last}^{t}(x,y) = f(M_{last}^{t}(x',  y')) \\  \forall \ x_i \ &\leq x' \leq x_i + w_i-1, \\ y_i \ &\leq y' \leq y_i + h_i-1, \\
t &\in {0, 1, \ldots, T-1}
\end{align*}

\noindent where $f(.)$ is an interpolation function which resizes the activation subset of size $w_i \times h_i$ to the size $L \times L$. In the above equation, $x, y \in \{ 0, \ldots, L-1\}$ and bi-linear interpolation is used.
After obtaining the reconstructed feature maps $\hat{M}_{last}$, we forward propagate the activations into the fully connected layers and obtain the class scores. An illustration of this step is presented in Figure $\ref{fig:activation_masking_cropped}$.

A limitation of the above approach is related to the fact that interpolating from a smaller subset to the larger grid will introduce interpolation artifacts into the reconstructed feature maps. In order to mitigate the effects of the artifacts, we limit the localization candidates to boxes with  $ L-1 \leq w_i \leq L $ and $L-1 \leq h_i \leq L$. From this limited corpus of localization candidates, we generate the corresponding $\hat{M}_{last}$ and consequently the object class scores, and choose the candidate with the highest class score. With the resultant localization candidate box $b_r$, we backproject onto the image space by cropping:

\begin{equation}
\begin{split}
x_{crop} &= \frac{x_r}{L}{W}, \
y_{crop} = \frac{y_r}{L}{H} \\
w_{crop} &= \frac{w_r}{L}{W}, \
h_{crop} = \frac{h_r}{L}{H}
\end{split}
\end{equation}

\begin{align*}
I_{crop}(x,y) = I(x+x_{crop}, y+y_{crop}) \ \forall \  0 \leq& \ x < w_{crop} \\ 0 \leq& \ y < h_{crop}
\end{align*}

\noindent where $x, y$ indicate pixel locations, and $W$ and $H$ are width and height of the test image respectively. We then repeat the above described localization process on $I_{crop}$ till a predetermined number of iterations. A visual example of progress in the iterative process is shown in Figure $\ref{fig:localization_iterations}$.

\subsection{Search Strategy}
\begin{figure}
\begin{center}
\includegraphics[width=0.5\columnwidth]{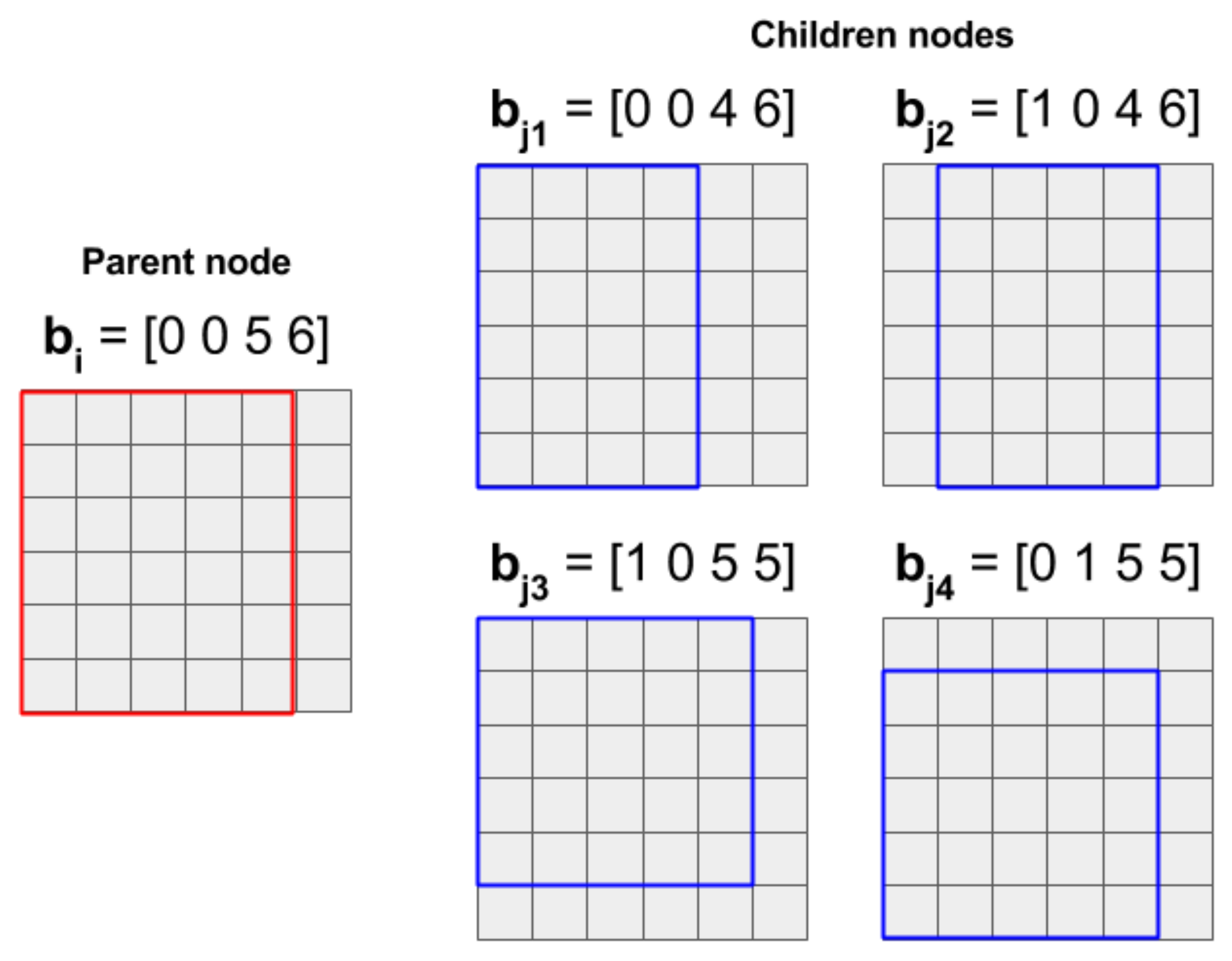}
\end{center}
   \caption{An example of a parent node (represented in red) and it's children nodes (represented in blue) displayed on a $6\times6$ grid, as is the case for the Alexnet $M_{5}$ feature maps. }
\label{fig:children}
\end{figure}

The localization strategy can be visualized as traversing down a search-tree where each node corresponds to a localization candidate $b_i$. The root node of such a tree would be $b_{0} = [0, 0, L, L]$. The children of a node $b_i$ in the tree would be the candidates $\{b_j\}$ which lie within sub-grid corresponding to $b_i$ and whose parameters $\{w_j\}$ and $\{h_j\}$ satisfy the below conditions:

\begin{equation}
w_i-1 \le w_j \le w_i, \ h_i-1 \le h_j \le h_i
\label{eq:children}
\end{equation}
We consider children nodes whose width or height values, but not both of them differ from the parent node by $1$. This restriction is put in place so that we are minimally modifying the feature map activations for discriminating amongst candidates. An example of a parent node $b_i$ and the corresponding  children node set  $\{b_j\}$ is shown in Figure $\ref{fig:children}$.

During traversal, the child candidate with the highest  score for the class $C$  is selected. This approach is a greedy search strategy where we follow one path from the root node to a leaf node which represents the finest localization, and is susceptible to arrival at a locally optimal solution. Alternatively, we could evaluate all the nodes in the entire search-tree and could come up with the localization candidate with the highest score for class $C$. However, this would be computationally prohibitive.
\begin{figure*}
\begin{center}
\includegraphics[width=\columnwidth]{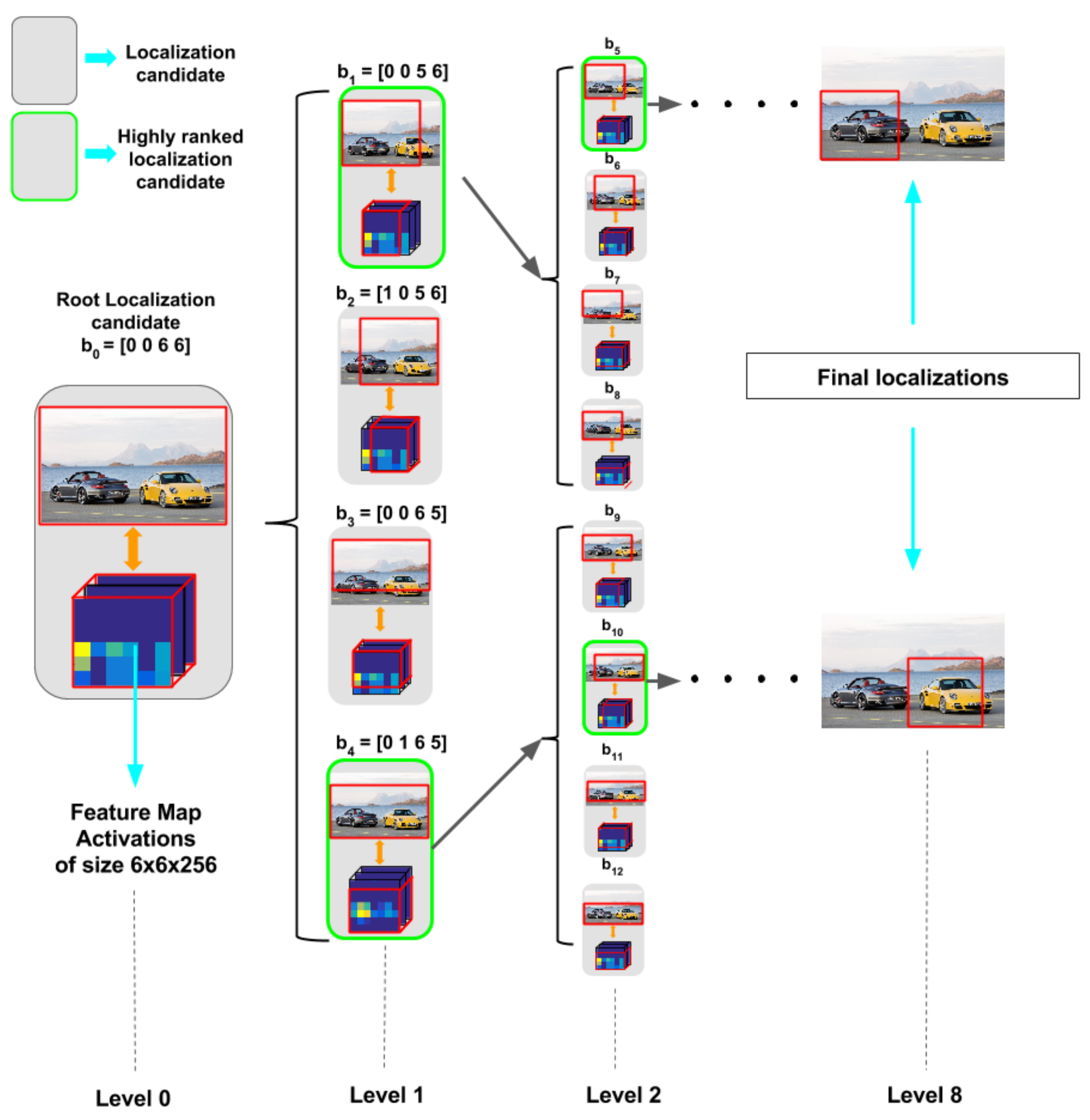}
\end{center}
   \caption{A visual example of beam-search strategy to navigate the search tree amongst localization candidates. In this specific case, the class $C$ is `car', $M$ is set to $2$ and $L$ is $6$.}
\label{fig:beamsearch}
\end{figure*}

To address this, we use the widely known beam-search \cite{rubin1977locus} strategy. At each level of the search-tree we generate sets of children nodes from the current set of localization candidates using Equation $\ref{eq:children}$. We then rank them according to the scores for class $C$. Only the top $M$ candidates are pursued for further evaluation. An illustration is presented in Figure $\ref{fig:beamsearch}$. In the Figure, we show an example where the two highest candidates are chosen at each level. The children nodes of these candidates are evaluated and ranked. We traverse a total of $H$ levels.  This approach helps us achieve a balance between keeping the number of computations to be tractable and avoiding greedy decisions. An additional advantage is the ability to localize multiple instances of the same class as the beam-search increases the set of localization candidates that are evaluated when compared to the greedy search strategy. Regions in the image corresponding to top-ranked candidates from each level are spatially sum-pooled using candidates scores to generate a heat-map. The heat-map is then threshold-ed. Bounding rectangles for the resulting binary blobs are extracted. The bounding rectangles are presented as detection results of our method. The average value of the heat-map values enclosed within detection boxes are assigned as the score of the boxes.
In our experiments, we have set the value of $M$ as 8 and $H$ in the search tree as 10 for all data-sets. Heat-map thresholds for each class were determined by evaluation on a small validation sub-set from the training set.

\section{Experiments}\label{section:Expertiments}

\subsection{Data-sets and Network training}
We evaluate our localization method on two large image data-sets, the PASCAL VOC 2007 \cite{pascal-voc-2007}, 2012 and the MS COCO. The VOC 2012 data-set has labels for 20 object categories and contains 5717 training images, 5823 validation images and 10991 test images. VOC 2007 shares the same class-labels with 2501 training images, 2510 validation images and 4952 test images. For the MS COCO data-set, there are 80000 images for training and 40504 images for validation with 80 object classes being present. These data-sets contain both image-level labels and object location annotations. For weak supervision we use the image-level labels from the training set to train classification networks and use the location annotations in the test and validation sets for evaluation.

\begin{figure*}
\begin{center}
\includegraphics[width=\columnwidth]{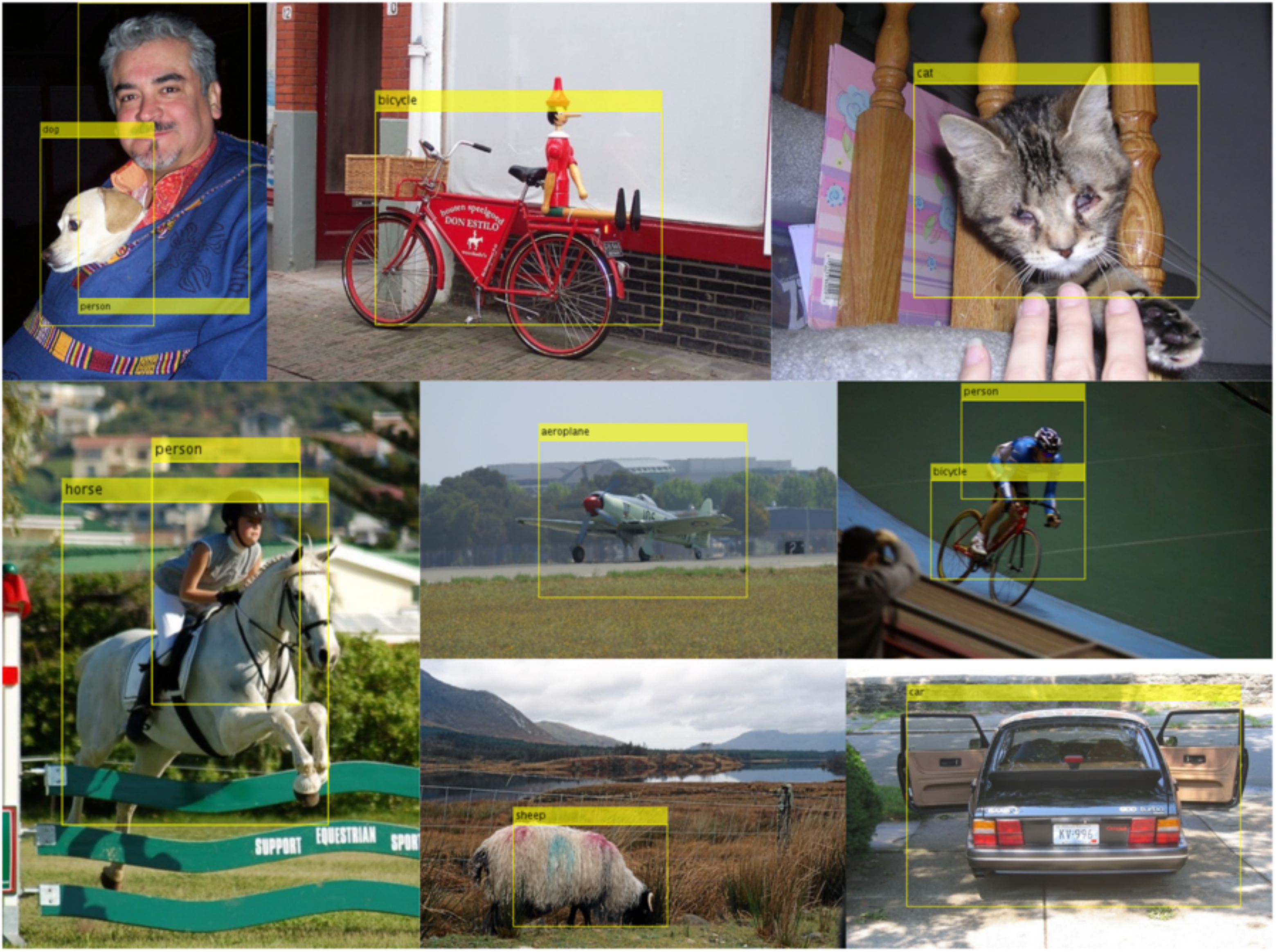}
\end{center}
\caption{Visual sample results from the proposed method for Pascal VOC 2007 test set. Yellow rectangles overlaid on the images represent location and extent predictions. The locations of objects in the shown images are accurately estimated. Considering that only image-level labels are used for training, extent estimations are a challenging problem in this setting.}
\label{fig:visualresults}
\end{figure*}

We fine-tune the original VGG-16 and Alexnet networks (trained on Imagenet) by re-training the final fully connected layer for the VOC 2007, 2012 and MS COCO data-sets. We set the learning rate parameter to 0.001 which we decrease by a factor 10 for every 20000 training batches. Each training batch consists of   50 samples and the network was trained with 400000 batches. In order to balance the data-sets with respect to number of samples per class, we oversampled training samples from under-represented classes. We generate additional samples by a combination of adding white gaussian noise and random rotations in the $\pm$ 30${}^\circ$  range. We use Caffe \cite{jia2014caffe} as our software platform for training and deploying classification networks on an NVIDIA TITAN X Desktop GPU.

\subsection{Metrics}
To compare results with the state-of-the-art in weakly supervised localization methods, we use the localization metric suggested by \cite{oquabobject}. From the class-specific heat-maps generated by our localization, we extract the region of maximal response. If the center location of the maximal response lies within the ground-truth bounding box of an object of the same class, we label the location prediction as correct. If not, the false positive count is increased as the background was assigned to the class, and the false negative count is increased because object was not detected. The maximal value of the heat-map is assigned as confidence of the localization. The confidence score is then used to rank localizations and associated precision-recall (p-r) curves are generated for each object class. The p-r curves are characterized by an estimate of the area under the curve, which is termed as the Average Precision (AP). The AP score can vary from 0 to 100. An $AP$ score of 100 signifies that all true positives were localized and no false positives were assigned scores. The $AP$ scores for all classes are averaged to derive the Mean Average Precision (mAP), which presents a summarized score for the entire test set. This evaluation metric differs from the traditional Intersection-over-Union ($IoU$) measures to determine bounding box quality w.r.t the ground-truth, as the extent of the localization is not captured.

In addition to the above metric, we are interested in measuring how effective our method is in capturing the extent of the object of interest. We calculate the standard average precision for our detection results, where true positives are determined when intersection over union (IoU) between the predicted bounding boxes and the corresponding ground-truth box of the same class exceeds 0.5.
\subsection{Results}


\begin{table}[]
\tiny
\centering
\begin{tabular}{|L|L|L|L|L|L|L|L|}
\hline
 & \multicolumn{2}{c|}{\textbf{Image Classification}} & \multicolumn{5}{c|}{\textbf{Localization}} \\ \cline{2-8}
 & \Longstack{Proposed Method + VGG-16} & \Longstack{Proposed Method + Alexnet} & \Longstack{Proposed Method + VGG-16} & \Longstack{Proposed Method + Alexnet} & \Longstack{Oquab et. al. \cite{oquabobject}} & \begin{tabular}[c]{@{}c@{}}RCNN$^{*}$\\ \cite{girshick2014rich}\end{tabular} & \begin{tabular}[c]{@{}c@{}}Fast-RCNN$^{*}$\\ \cite{girshickICCV15fastrcnn}\end{tabular} \\ \hline
airplane & 93.0 & 92.0 & 90.1 & 90.0 & 90.3 & 92.0 & 79.2 \\ \hline
bike & 89.7 & 82.9 & 86.4 & 81.2 & 77.4 & 80.8 & 74.7 \\ \hline
bird & 91.4 & 87.2 & 86.4 & 81.2 & 77.4 & 80.8 & 74.7 \\ \hline
boat & 89.6 & 83.8 & 77.6 & 82.2 & 79.2 & 73.0 & 65.8 \\ \hline
bottle & 69.5 & 54.1 & 56.8 & 47.5 & 41.1 & 49.9 & 39.4 \\ \hline
bus & 90.9 & 87.3 & 90.3 & 86.7 & 87.8 & 86.8 & 82.3 \\ \hline
car & 81.6 & 74.5 & 68.3 & 64.9 & 66.4 & 77.7 & 64.8 \\ \hline
cat & 92.0 & 87.0 & 89.9 & 85.7 & 91.0 & 87.6 & 85.7 \\ \hline
chair & 69.3 & 56.4 & 54.7 & 53.9 & 47.3 & 50.4 & 54.5 \\ \hline
cow & 88.9 & 76.7 & 86.8 & 75.8 & 83.7 & 72.1 & 77.2 \\ \hline
\Longstack{dining table} & 80.2 & 71.1 & 66.4 & 67.9 & 55.1 & 57.6 & 58.8 \\ \hline
dog & 90.4 & 83.5 & 88.5 & 82.2 & 88.8 & 82.9 & 85.1 \\ \hline
horse & 90.0 & 85.5 & 89.0 & 84.1 & 93.6 & 79.1 & 86.1 \\ \hline
motorbike & 90.0 & 84.3 & 88.1 & 83.4 & 85.2 & 89.8 & 80.5 \\ \hline
person & 91.6 & 88.1 & 78.5 & 83.9 & 87.4 & 88.1 & 76.6 \\ \hline
plant & 85.5 & 80.1 & 64.1 & 71.7 & 43.5 & 56.1 & 46.7 \\ \hline
sheep & 90.4 & 83.5 & 90.0 & 83.1 & 86.2 & 83.5 & 79.5 \\ \hline
sofa & 75.5 & 64.5 & 67.0 & 63.7 & 50.8 & 50.1 & 68.3 \\ \hline
train & 91.4 & 90.8 & 89.9 & 89.4 & 86.8 & 82.0 & 85.0 \\ \hline
tv & 89.6 & 81.4 & 82.6 & 78.2 & 66.5 & 76.6 & 60.0 \\ \hline
\textbf{mAP} & 86.5 & 79.8 & \textbf{79.7} & \textbf{77.1} & 74.5 & 74.8 & 71.3 \\ \hline
\end{tabular}
\caption{Comparison of Image classification and Object Localization scores on the PASCAL VOC 2012 $\textit{validation}$ set. For computing localization scores, responses are labeled as correct when the maximal responses fall within a ground-truth bounding box of the same class. False negatives are counted when no responses overlap with the ground-truth annotations. The class scores of the associated image-level classification are used to rank the responses and generate average precision scores. * RCNN and Fast-RCNN are trained for object detection with object-level bounding box data. We use the most confident bounding box per class in every image for evaluation.}
\label{table:localizationVOC2012}

\end{table}

 For obtaining localization results, we fine-tuned the networks using training samples from the $\textit{train}$ set of PASCAL VOC 2012 data-set and tested the trained networks on the $\textit{validation}$ set. As we use the class-scores from the classifiers to drive our localization strategy, good classification performance is essential for robust object localization. We present the classification performance on the PASCAL VOC 2012 $\textit{validation}$ set in Table $\ref{table:localizationVOC2012}$. The VGG-16 network provides improved classification with respect to Alexnet and a consequent improvement can be seen in the localization scores as well.

In Table $\ref{table:localizationVOC2012}$, we also compare the localization results of our method with respect to recent state-of-the-art weakly supervised localization methods on the PASCAL VOC 2012 $\textit{validation}$ set. We achieve a significant improvement of 5 mAP over the  localization performance of Oquab et.al \cite{oquabobject}. We also compare against the RCNN \cite{girshick2014rich} and Fast RCNN \cite{girshickICCV15fastrcnn} detectors which are trained with object-level bounding boxes. Similar to the way \cite{oquabobject} evaluates \cite{girshick2014rich}, we select the most confident bounding box proposal per class per image for evaluation. Since deep neural networks are the state-of-the-art in object detection and localization tasks, we have compared with CNN-based methods.

We summarize the localization results for the much larger MS COCO \textit{validation} data-set in Table $\ref{table:localizationMSCOCO}$. Inspite of having weaker classification performance (54.1 mAP vs 62.8 mAP) than the network used by \cite{oquabobject}, we are able to produce stronger localization performance by a large margin of 8 mAP. This is a significant improvement in performance over the state-of-the-art method. This is mainly because the proposed method actively seeks out image regions triggering higher classification scores for the class of interest. This form of active learning, where the localizing algorithm is the weak learner and the classifier is the strong teacher, lends us an advantage when trying to localize objects in complex scenes where multiple objects can exist in varying mutual configurations. This is also observed for the PASCAL VOC 2012 data-set. The fine-tuned VGG-16 and Alexnet networks produce classification performance scores of  74.3 mAP and 82.4 mAP respectively on the $\textit{test}$ set, where as the network used by \cite{oquabobject} is scored at 86.3 mAP. As noted before, the proposed method outperforms competing methods on the localization task.

We have provided results on object bounding box detection for the PASCAL VOC 2007 $\textit{test}$ set in Table $\ref{table:detectionscores}$. We fine-tuned our network on the VOC 2007 $\textit{train}$ and the $\textit{validation}$ set, where 10\% of this joint group of images was set aside for parameter tuning, and provide test results on the $\textit{test}$ set. We are comparable in performance with respect to other state-of-the-art weakly supervised methods \cite{cinbis2014multi}, \cite{bilen2015weakly} and \cite{wang2014weakly}.  Examples of visual results for object detection are provided in Figure $\ref{fig:visualresults}$. We have also compared with the detection performance of the proposed method with results from \cite{oquabobject} on the VOC 2012 validation set, where we trained the classifier on the $\textit{train}$ set. We demonstrate a marked improvement in mAP scores.

Re-scoring the class likelihood scores using co-occurrence information referenced in equation $\ref{eq:rescoring}$ contributes to an improvement of 1.2 with the VGG-16 network in classification mAP score and 0.8 localization mAP score from Table $\ref{table:localizationVOC2012}$.

\begin{table}[]
\small
\centering
\begin{tabular}{|l|c|}
\hline
\textbf{Method} & \multicolumn{1}{l|}{Localization score (mAP)} \\ \hline
Oquab et. al. \cite{oquabobject} & 41.2 \\ \hline
Proposed Method + VGG-16 & \textbf{49.2} \\ \hline
\end{tabular}
\caption{Comparison of localization and classification mAP scores for the MS COCO validation set.}
\label{table:localizationMSCOCO}
\end{table}


\begin{table}[]
\footnotesize
\setlength{\tabcolsep}{2pt}
\centering
\begin{tabular}{|l|c|}
\hline
\textbf{Method}                         & \multicolumn{1}{l|}{\textbf{mAP}} \\ \hline
Multi-fold MIL \cite{cinbis2014multi} & 22.4                                   \\ \hline
Bilen et. al. \cite{bilen2015weakly}  & \textbf{27.7}                                   \\ \hline
LCL-kmeans \cite{wang2014weakly}      & 26.9                                   \\ \hline
Proposed Method + VGG-16                & 25.7                                   \\ \hline
\end{tabular}
\caption{Comparison of mean average precision scores for Object Detection task on the PASCAL VOC 2007 test set.}
\label{table:detectionscores}
\end{table}

\begin{table}[]
\centering

\label{my-label}
\begin{tabular}{|l|c|}
\hline
\textbf{Method}                                                     & \multicolumn{1}{l|}{\textbf{mAP}} \\ \hline
Proposed Method + VGG 16                                            & \textbf{26.5}                     \\ \hline
Oquab et. al. \cite{oquabobject} + Selective Search \cite{uijlings2013selective} & 11.7                              \\ \hline
\end{tabular}
\caption{Comparison of mean average precision scores for Object Detection
task on the PASCAL VOC 2012 validation set.}
\label{table:2012detectionscores}
\end{table}

\section{Discussion and Conclusions}\label{section:Discussion}
The proposed method requires 2.6 sec to localize an object on an image on machine with a 2.3 GHz CPU with a NVIDIA TITAN X desktop GPU. Compared to region proposal-based detection methods like RCNN which take around 20 seconds to detect objects, we achieve a significant reduction in localization time.

 As can be seen from Table $\ref{table:localizationVOC2012}$, an improvement in the classification performance (e.g. from Alexnet to VGG-16) directly leads to an improvement in the localization performance. As the state-of-the-art of the classification CNNs improves, we can expect a similar improvement in localization performance from our proposed method.

In summary, this method directly leverages feature map activations for object localization. This work uses the spatial and semantic information encoded in the convolutional layers and we have explored methods to utilize activations in  the last convolutional layer. It would be interesting to see the improvements that could be derived by combining coarser semantic and finer localization information in earlier convolutional layers as well. Another direction to explore would be combining fast super-pixel segmentation and localization candidates from proposed method to improve detection performance.

The proposed method relies on weak supervision, with networks trained for image classification being used for localizing objects in test images with complex scenes and hence opens up possibilities for extending object localization to new object categories and image modalities without requiring expensive object-level annotations.

\clearpage

\bibliographystyle{splncs03}

\end{document}